# Inverse Kinematics and Dexterous Workspace Formulation for 2-Segment Continuum Robots with Inextensible Segments

Yifan Wang, Zhonghao Wu, Longfei Wang, Bo Feng and Kai Xu*, *Member, IEEE*

*Abstract*—The inverse kinematics (IK) problem of continuum robots has been investigated in depth in the past decades. Under the constant-curvature bending assumption, closed-form IK solution has been obtained for continuum robots with variable segment lengths. Attempting to close the gap towards a complete solution, this paper presents an efficient solution for the IK problem of 2-segment continuum robots with one or two inextensible segments (a.k.a, constant segment lengths). Via representing the robot's shape as piecewise line segments, the configuration variables are separated from the IK formulation such that solving a one-variable nonlinear equation leads to the solution of the entire IK problem. Furthermore, an in-depth investigation of the boundaries of the dexterous workspace of the end effector caused by the configuration variables limits as well as the angular velocity singularities of the continuum robots was established. This dexterous workspace formulation, which is derived for the first time to the best of the authors' knowledge, is particularly useful to find the closest orientation to a target pose when the target orientation is out of the dexterous workspace. In the comparative simulation studies between the proposed method and the Jacobian-based IK method involving 500,000 cases, the proposed variable separation method solved 100% of the IK problems with much higher computational efficiency.

## I. INTRODUCTION

Continuum robots demonstrate potentials for applications in industrial inspection [1], rescue [2], and healthcare [3] due to their advantages, such as dexterity in confined spaces, inherent compliance, and structural compactness.

Kinematics modeling of multi-segment continuum robots usually adopts a common assumption of constant curvature bending [4]. This approach has been verified analytically and experimentally [5, 6]. It is widely used due to its analytic formulation of the forward kinematics. The inverse kinematics (IK) problem, on the other hand, is not straightforward given the fact that segment bending changes the position and orientation of the end effector at the same time to introduce strong coupling between the end effector's position and orientation. The existence of a closed-form IK solution depends on the robot's specific structure and is valid only for the ones with variable-length segments as in [7-9] and the ones with an inverted dual continuum mechanism and inextensible continuum segments as in [10]. When the closed-form IK solution is not available, numerical approaches are often adopted, e.g., the Jacobian-based methods [5, 11-13]. However, the Jacobian-based methods do not always converge to a solution if improper initial guesses of the configuration variables were adopted. Besides, its computational requirement is also relatively high. Hence, an efficient solution to the IK problem of continuum robots with inextensible bending segments is still in great need, since inextensible continuum segments can be much more reliably fabricated in practice.

The challenge of solving the IK problem of a continuum robot comes from the coupling between the position and orientation of its end effector when continuum segments are bent. Due to this strong coupling, all the existing Jacobian-based numerical methods solve all configuration variables simultaneously from the multi-dimensional IK formulation.

In this paper, an efficient Variable Separation IK (VS-IK) method for 2-segment constant-curvature continuum robots with one or two inextensible bending segments is hence presented. By representing the robot shape as piecewise line segments and utilizing the formulation of the line segment geometry, one key configuration variable is separated from the IK formulation. And a one-dimensional nonlinear equation is solved subsequently to obtain one and then all the configuration variable values. The computational efficiency is greatly improved since the to-be-solved equation is only one-dimensional. The conducted simulation studies showed that the VS-IK method significantly outperformed the Jacobian-based method in terms of computation time, number of iterations, and success rate.

Furthermore, the boundaries of the dexterous workspace (i.e., reachable orientations of the end effector at a target position) are categorized into two types. One type of the boundaries caused by the configuration variables limits is analytically formulated, while the other type of the boundaries that occur at singularities of the angular velocity is formulated implicitly. To the best of the authors' knowledge, this categorization is the first one to analyze the dexterous workspace of a continuum robot. The existing studies evaluate the dexterous workspace of the continuum robots via exhausting numerical methods [14-16]. Based on the presented investigation, handling the target poses with unreachable orientations becomes much easier.

*This work was supported in part by the National Key R&D Program of China (Grant No. 2019YFC0118003, Grant No. 2017YFC0110800 and Grant No. 2019YFC0118004), in part by the Foundation of National Facility for Translational Medicine (Shanghai) (Grant No. TMSK-2021-505), in part by the National Natural Science Foundation of China (Grant No. 51722507).

Yifan Wang, Longfei Wang and Kai Xu are with the State Key Laboratory of Mechanical System and Vibration, School of Mechanical Engineering, Shanghai Jiao Tong University, Shanghai, 200240, China (e-mails: fan_tasy@sjtu.edu.cn, longfei.wang@sjtu.edu.cn and k.xu@sjtu.edu.cn; corresponding author: Kai Xu).

Zhonghao Wu is with the RII Lab (Laboratory of Robotics Innovation and Intervention), UM-SJTU Joint Institute, Shanghai Jiao Tong University, Shanghai, 200240, China (e-mail: zhonghao.wu@sjtu.edu.cn).

Bo Feng is with Department of Surgery, Affiliated Ruijin Hospital, Shanghai Jiao Tong University, Shanghai, China, 200025 (e-mail: fengbo2022@163.com).

The proposed VS-IK method is shown applicable to continuum robots with one or two inextensible segments, even when the two segments are not directly connected.

This paper is organized as follows. Section II summarizes the kinematics of the 2-segment continuum robot. Next, the geometry-based VS-IK method is elaborated in Section III. In Section IV, the analytic formulation of the dexterous workspace is detailed, and the IK solution that considers the unreachable orientations is described. Simulation study of the VS-IK method compared to the Jacobian-based method is reported in Section V. The conclusion and future works are summarized in Section VI.

## II. Kinematics Nomenclature and Coordinates

The continuum robot under investigation consists of two bending segments, each with 2 bending degrees of freedom (DoFs), a rigid straight base stem, a rigid straight middle stem, and an end effector. A practical embodiment of the continuum robot is as in [17]. The robot driven by an actuation unit can be deployed into a working cavity through a feed channel, and can work in both partially or fully inserted configurations.

When the 1st bending segment is partially inserted into the cavity, this inserted portion is equivalent to a segment with variable length. Hence, the 1st segment possesses 2-DoF bending and 1-DoF length varying, while the 2nd segment possesses 2-DoF bending with a constant length. The actuation unit provides 1-DoF rotation about the neutral axis of the 1st segment. This configuration only has the 2nd segment as an inextensible segment, as shown in Fig. 1(a.1), and is therefore referred to as configuration CI-1.

When the 1st bending segment is fully inserted, a configuration transition occurs such that the base stem now introduces a translation along the axis of the feed channel. Hence, the two bending segment each possesses 2-DoF bending, and the actuation unit provides 1-DoF rotation about the axis of the base stem. This configuration has two inextensible segments, as shown in Fig. 1(a.2), and is referred to as configuration CI-2.

### A. Nomenclature and Coordinate

The coordinate attachment for the entire robot are shown in Fig. 1(a), while the coordinate attachment for the $t^{th}$ segment in Fig. 1(b). The coordinates are defined as follows, and the nomenclature is defined in Table I.

- Base coordinate $\{tb\} \equiv \{\hat{\mathbf{x}}_{tb}, \hat{\mathbf{y}}_{tb}, \hat{\mathbf{z}}_{tb}\}$ is attached to the center of the base cross section of the $t^{th}$ segment with $\hat{\mathbf{z}}_{tb}$ perpendicular to the base cross section.

- End coordinate $\{te\} \equiv \{\hat{\mathbf{x}}_{te}, \hat{\mathbf{y}}_{te}, \hat{\mathbf{z}}_{te}\}$ is attached to the center of the end cross section of the $t^{th}$ segment. $\hat{\mathbf{z}}_{te}$ is perpendicular to the end cross section, and $\hat{\mathbf{x}}_{te}$ points to the same segment surface region as $\hat{\mathbf{x}}_{tb}$ such that the segment does not undergo twisting.

- Bending plane coordinate 1 $\{t1\} \equiv \{\hat{\mathbf{x}}_{t1}, \hat{\mathbf{y}}_{t1}, \hat{\mathbf{z}}_{t1}\}$ is defined such that the $t^{th}$ continuum segment bends in its XY plane, and its origin coincides with that of $\{te\}$. The XY plane of $\{t1\}$ is referred to as the bending plane.

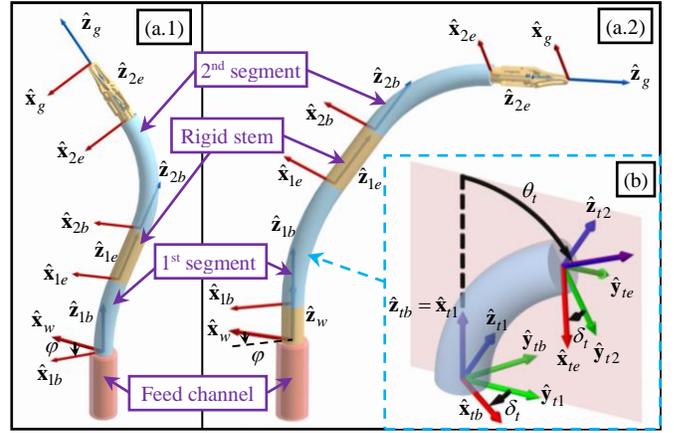

Fig. 1 Nomenclature, coordinates and configurations of the continuum robot: (a.1) configuration CI-1; (a.2) configuration CI-2; (b) the $t^{th}$ segment.

TABLE I. Nomenclatures Used in This Paper

| Symbol | Representation |
|---|---|
| $t$ | Index of the segments. $t = 1, 2$. |
| $L_t, L_{t0}$ | Inserted length and full length of the $t^{th}$ segment. $0 \leq L_t \leq L_{t0}$. |
| $L_r$ | Length of the straight rigid stem. |
| $L_s, L_{s+}$ | Inserted length of the base stem and its upper limit. $0 \leq L_s \leq L_{s+}$. |
| $L_g$ | Length of the end effector. |
| $\theta_t, \theta_{t+}$ | $\theta_t$ is the segment bending angle, namely the rotation angle from $\hat{\mathbf{x}}_{t1}$ about $\hat{\mathbf{z}}_{t1}$ to $\hat{\mathbf{x}}_{t2}$; $\theta_{t+}$ is the upper limit of $\theta_t$. $0 \leq \theta_t \leq \theta_{t+}$. |
| $\delta_t$ | Rotation angle from $\hat{\mathbf{y}}_{t1}$ about $\hat{\mathbf{z}}_{tb}$ to $\hat{\mathbf{x}}_{tb}$, indicating segment bending direction. |
| $\varphi$ | Rotation angle provided by the actuation unit to the 1st segment or the base stem. |
| $r_t, r_{t-}$ | $r_t$ is the bending radius of the $t^{th}$ segment, and $r_{t-}$ is the lower limit of $r_t$. $r_{t-} \leq r_t < \infty$. |

- Bending plane coordinate 2 $\{t2\} \equiv \{\hat{\mathbf{x}}_{t2}, \hat{\mathbf{y}}_{t2}, \hat{\mathbf{z}}_{t2}\}$ is obtained from $\{t1\}$ via a rotation about $\hat{\mathbf{z}}_{t1}$ for an angle $\theta_t$ so that $\hat{\mathbf{x}}_{t2}$ is aligned with $\hat{\mathbf{z}}_{te}$.

- End effector coordinate $\{g\} \equiv \{\hat{\mathbf{x}}_g, \hat{\mathbf{y}}_g, \hat{\mathbf{z}}_g\}$ is attached to the tip of the end effector, obtained by moving $\{2e\}$ for the length $L_g$ along $\hat{\mathbf{z}}_{2e}$.

- World coordinate $\{w\} \equiv \{\hat{\mathbf{x}}_w, \hat{\mathbf{y}}_w, \hat{\mathbf{z}}_w\}$ is fixed to the feed channel with $\hat{\mathbf{z}}_w$ aligned with the channel axis. It should be noticed that $\{1b\}$ is obtained from $\{w\}$ via a rotation about $\hat{\mathbf{z}}_w$ for an angle $\varphi$.

According to the nomenclature, the configuration variables for CI-1 are $\varphi, \theta_1, L_1, \delta_1, \theta_2, \delta_2$, while the configuration variables for CI-2 are $L_s, \varphi, \theta_1, \delta_1, \theta_2, \delta_2$.

### B. Forward Kinematics

The position and orientation of $\{te\}$ for the $t^{th}$ segment are given by (1) and (2).

$$^{tb}\mathbf{p}_{te} = \frac{L_t}{\theta_t}\left[\cos\delta_t(1-\cos\theta_t) \quad \sin\delta_t(1-\cos\theta_t) \quad \sin\theta_t\right]^T, \quad (1)$$

$$^{tb}\mathbf{R}_{te} = \text{Rot}(\hat{\mathbf{z}}, -\delta_t)\text{Rot}(\hat{\mathbf{y}}, \theta_t)\text{Rot}(\hat{\mathbf{z}}, \delta_t), \quad (2)$$

where $^{tb}\mathbf{p}_{te} = [0\ 0\ L_t]^T$ when $\theta_t = 0$, and $\text{Rot}(\hat{\mathbf{m}}, \alpha)$ represents the rotation matrix about the axis $\hat{\mathbf{m}}$ by an angle $\alpha$.

The complete forward kinematics of the continuum robot can be referred to [17].

## III. Variable Separation Inverse Kinematics

The VS-IK method for the configurations CI-2 and CI-1 are discussed in Section III.A and Section III.B, respectively.

### A. Inverse Kinematics for the Configuration CI-2

To separate the configuration variables, the robot is characterized using piecewise line segments, as shown in Fig. 2 (a). A constant-curvature continuum segment is represented by two line segments that are tangent to the arc at its two ends, as shown in Fig. 2 (b). Then, the shape of the entire continuum robot is represented by three consecutive line segments. The intersections of the line segments are $\mathbf{p}_1$ and $\mathbf{p}_2$, respectively.

The line segment length of the $t^{th}$ segment is given by (3).

$$l_t = L_t \tan(\theta_t/2)/\theta_t, \quad (3)$$

where $l_t = L_t/2$ when $\theta_t = 0$. Note that $l_t$ approaches infinity for $\theta_t = \pi$. In the following content, the maximum bending angle $\theta_{t+}$ is assumed to be less than $\pi$, which is practically adopted for most continuum robot designs.

The pose of the end effector is represented by (4).

$$^w\mathbf{T}_g = \begin{bmatrix} ^w\mathbf{R}_g & ^w\mathbf{p}_g \\ \mathbf{0}_{1\times 3} & 1 \end{bmatrix} = \begin{bmatrix} \mathbf{n} & \mathbf{s} & \mathbf{a} & ^w\mathbf{p}_g \\ & \mathbf{0}_{1\times 3} & & 1 \end{bmatrix}, \quad (4)$$

where $^w\mathbf{p}_g = [p_x\ p_y\ p_z]^T$ denotes the position of the end effector, while $\mathbf{n}$, $\mathbf{s}$, and $\mathbf{a}$ are unit vectors.

In the line segment representation, the position of $\mathbf{p}_1$ is expressed from the origin of $\{w\}$, while the position of $\mathbf{p}_2$ is expressed backward from $^w\mathbf{p}_g$ as follows.

$$\mathbf{p}_1 = \begin{bmatrix} 0 & 0 & l_1 + L_s \end{bmatrix}^T, \quad (5)$$

$$\mathbf{p}_2 = {}^w\mathbf{p}_g - L_{2g}\mathbf{a} = \begin{bmatrix} p_x - L_{2g}a_x & p_y - L_{2g}a_y & p_z - L_{2g}a_z \end{bmatrix}^T, \quad (6)$$

where $L_{2g} = l_2 + L_g$.

And it follows that

$$\|\mathbf{p}_2 - \mathbf{p}_1\|^2 = L_{1r2}^2, \quad (7)$$

where $L_{1r2} = l_1 + L_r + l_2$.

Then, $\theta_t$ is expressed using the inner products of the vectors available in the line segment representation as follows.

$$\cos\theta_1 = (\mathbf{p}_2 - \mathbf{p}_1)^T \hat{\mathbf{z}}_w / L_{1r2} = (\mathbf{p}_2|_z - \mathbf{p}_1|_z)/L_{1r2}, \quad (8)$$

$$\cos\theta_2 = (\mathbf{p}_2 - \mathbf{p}_1)^T \mathbf{a} / L_{1r2}. \quad (9)$$

Expanding (8) gives an expression of $L_s$ with respect to $l_1$, $l_2$ and $\theta_1$, as in (10). Substituting (10) into (7) yields (11).

$$L_s = p_z - L_{2g}a_z - L_{1r2}\cos\theta_1 - l_1, \quad (10)$$

$$(\cos^2\theta_1 - a_z^2)l_2^2 + (l_1 + L_r)^2(\cos^2\theta_1 - 1) + b_1 \\ + 2[(l_1 + L_r)(\cos^2\theta_1 - 1) - b_2]l_2 = 0, \quad (11)$$

where $b_1 = p_x^2 + p_y^2$, and $b_2 = p_x a_x + p_y a_y$.

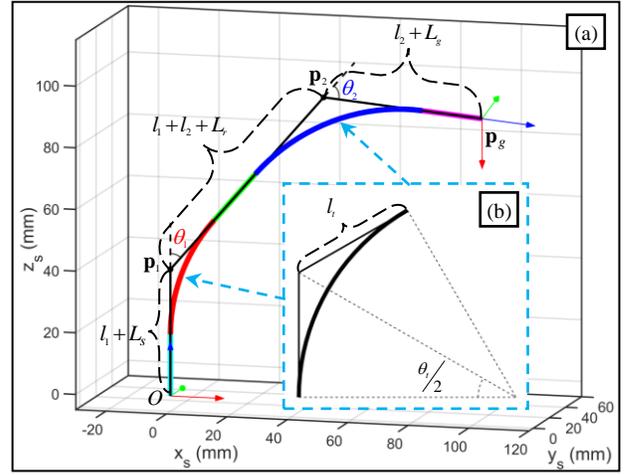

Fig. 2 Line segment representation for (a) the continuum robot under Configuration CI-2, and (b) a bending segment.

Substituting (10) into (9) gives an expression of $\cos\theta_2$ with respect to $l_1$, $l_2$ and $\theta_1$, as in (12). Then, substituting $\theta_2$ and $\tan(\theta_2/2)$ in (3) using (12) yields (13).

$$\cos\theta_2 = ({}^w\mathbf{p}_g^T\mathbf{a} + L_{2g}(a_z^2 - 1) - a_z(p_z - L_{1r2}\cos\theta_1))/L_{1r2}, \quad (12)$$

$$l_2 \arccos\left(\frac{b_3}{L_{1r2}} + a_z \cos\theta_1\right) \sqrt{\frac{(1 + a_z \cos\theta_1)L_{1r2} + b_3}{(1 - a_z \cos\theta_1)L_{1r2} - b_3}} = L_2, \quad (13)$$

where $b_3 = b_2 + (a_z^2 - 1)l_2$.

Since $l_1$ and $l_2$ are functions of $\theta_1$ as shown in (3) and (11), equation (13) only contains the configuration variable $\theta_1$, and can be solved using general nonlinear equation solving methods. Please note that equation (11) gives two possible solutions of $l_2$. While solving (13), one solution of $l_2$ from (11) is firstly used; if no solution of $\theta_1$ is found, the other solution of $l_2$ is used to continue the iteration.

After solving $\theta_1$, $l_1$ and $l_2$ are obtained from (3) and (11) respectively. Then, $L_s$ is obtained from (7) or (8), and $\theta_2$ is obtained from (9). The remaining configuration variables $\delta_1$, $\delta_2$ and $\varphi$ are solved as follows.

Using (5) and (6), $\mathbf{p}_1$ and $\mathbf{p}_2$ are obtained. Points $\mathbf{p}_1$, $\mathbf{p}_2$ and the origin of $\{w\}$ lie in the bending plane of the 1st segment, while $\mathbf{p}_1$, $\mathbf{p}_2$ and $\mathbf{p}_g$ lie in the bending plane of the 2nd segment. The bending direction of the 1st segment is calculated by:

$$\varphi - \delta_1 = \arctan 2(\mathbf{p}_2|_y, \mathbf{p}_2|_x). \quad (14)$$

The relative bending direction between the 1st and the 2nd segments is calculated by:

$$\delta_1 - \delta_2 = \text{sgn}\left((\mathbf{p}_1 \times \mathbf{p}_2)^T \mathbf{p}_g\right) \cdot \\ \arccos\left(\left(\frac{\hat{\mathbf{z}}_w \times \mathbf{p}_2}{\|\mathbf{p}_2\|}\right)^T \left(\frac{(\mathbf{p}_2 - \mathbf{p}_1) \times \mathbf{a}}{L_{1r2}}\right)\right), \quad (15)$$

where the sgn function indicates the direction from $\delta_1$ to $\delta_2$. The sgn function is zero when $(\mathbf{p}_1 \times \mathbf{p}_2)^T \mathbf{p}_g = 0$, indicating that both segments are bending to the same direction and $\delta_1 = \delta_2$.

The forward kinematics for the end effector orientation is given as follows:

$$\text{Rot}(\hat{\mathbf{z}}, \varphi)\,^{1b}\mathbf{R}_{1e}\,^{2b}\mathbf{R}_{2e} = \,^{w}\mathbf{R}_{g} \,. \tag{16}$$

Substituting (2) into (16) yields (17).

$$\begin{aligned}&\text{Rot}(\hat{\mathbf{z}},-\delta_2) = \\ &\,^{w}\mathbf{R}_g^T \text{Rot}(\hat{\mathbf{z}},\varphi-\delta_1)\text{Rot}(\hat{\mathbf{y}},\theta_1)\text{Rot}(\hat{\mathbf{z}},\delta_1-\delta_2)\text{Rot}(\hat{\mathbf{y}},\theta_2)\end{aligned}. \tag{17}$$

Using (14) and (15), the terms on the right side of (17) are all known, and $\delta_2$ can be calculated. Then, $\delta_1$ is obtained using (15), and $\varphi$ is obtained using (14).

In summary, the mapping from the piecewise line segments to the configuration under the constant curvature model is given by (8), (9), (14), (15), and (17).

### B. Inverse Kinematics for the Configuration CI-1

For the configuration CI-1, please note $L_s = 0$ and (7) only contains two variables ($l_1$ and $l_2$). Equation (7) gives (18).

$$l_1 = (c_1 l_2 + c_2)/(c_3 l_2 + c_4), \tag{18}$$

where the coefficients are

$$c_1 = -2(\,^w\mathbf{p}_g^T\mathbf{a} - L_g + L_r), \quad c_2 = \,^w\mathbf{p}_g^T(\,^w\mathbf{p}_g - 2L_g\mathbf{a}) + L_g^2 - L_r^2,$$
$$c_3 = 2(1-a_z), \qquad c_4 = 2(p_z - a_z L_g + L_r).$$

The expression for $\theta_t$ is again given by (8) and (9), with $L_s = 0$. Since $L_1$ is a configuration variable in the configuration CI-1, $l_1$ in (9) is substituted with (18), resulting in (19) that only contains the configuration variable $\theta_2$.

$$c_3(\cos\theta_2 + 1)l_2^2 + d_1 l_2 \cos\theta_2 + d_2 l_2 + d_3 \cos\theta_2 + d_4 = 0, \tag{19}$$

where the coefficients are

$$d_1 = c_1 + c_4 + c_3 L_r, \quad d_2 = c_4 + c_1 a_z - c_3(\,^w\mathbf{p}_g^T\mathbf{a} - L_g),$$
$$d_3 = c_2 + c_4 L_r, \qquad d_4 = c_2 a_z - c_4(\,^w\mathbf{p}_g^T\mathbf{a} - L_g).$$

Substituting $l_2$ with (3) into (19) leads to a nonlinear equation about $\theta_2$ that can be efficiently solved.

After solving $\theta_2$, $l_2$ is calculated by (3), and then $l_1$ is calculated by (18). Equation (8) is used to calculate $\theta_1$, and then (3) is used to calculate $L_1$. Finally, $\delta_1$, $\delta_2$ and $\varphi$ are calculated in the same way as that for the configuration CI-2.

## IV. FORMULATION OF THE DEXTEROUS WORKSPACE BOUNDARIES

There are two types of dexterous workspace boundaries: 1) the boundaries caused by configuration limits (type-I boundary), and 2) the boundaries that occur when the Jacobian of the angular velocity becomes singular at a fixed position (type-II boundary).

When the Jacobian-based method is used for an IK problem, a concerning issue is that when the method fails, it is usually uncertain whether the target pose is unreachable or the solution process fails to converge, even when the target position is within the translational workspace. Analytic formulation of the type-I boundaries of the dexterous workspace, which is firstly derived here to the best of the authors' knowledge, as well as the investigation of the type-II boundaries, greatly facilitates the IK solution, particularly when the target orientation is unreachable.

### A. Dexterous Workspace of the Configuration CI-1

The rotation of the end effector about its axis $\hat{\mathbf{z}}_g$ can be independently generated by changing $\varphi$ without affecting $\hat{\mathbf{z}}_g$. Therefore, the dexterous workspace only concerns about $\hat{\mathbf{z}}_g$ (the pointing direction), which is characterized by a unit spherical surface centered at the target point.

Since the configuration variables $\delta_1$ and $\delta_2$ are not limited, there exist two feasible configurations with mirrored shapes about a symmetry plane for a given end effector position, as illustrated in Fig. 3(a.1). The symmetry plane is defined by the target end effector position $^w\mathbf{p}_g$ and the $\hat{\mathbf{z}}_w$ axis of the world coordinates. The $\hat{\mathbf{z}}_g$ axis of the two mirrored poses are therefore symmetric about the plane, indicating the symmetry of the dexterous workspace with respect to the plane. Hence, it is sufficient to investigate the dexterous workspace by projecting half of the unit spherical surface onto the symmetry plane to form a concentric unit circular area. And the dexterous workspace can be described in the unit circular area, as shown in Fig. 3(a.2 and a.3). A symmetry plane coordinate $\{s\}$ is then defined as in Fig. 3, rotating $\{w\}$ around $\hat{\mathbf{z}}_w$ such that the XZ-plane of $\{s\}$ coincides with the symmetry plane. Please note that all equations derived in Section II still hold while expressed in $\{s\}$, with the target position $^w\mathbf{p}_g$ and the $\hat{\mathbf{z}}$ axis of the target orientation $\mathbf{a}$ represented by $^s\mathbf{p}_g = [p_{sx}\, p_{sy}\, p_{sz}]^T$ and $^s\mathbf{a} = [a_{sx}\, a_{sy}\, a_{sz}]^T$. They are obtained as follows:

$$^s\mathbf{p}_g = \text{Rot}(\hat{\mathbf{z}},-\gamma)\mathbf{p}_g, \quad ^s\mathbf{a} = \text{Rot}(\hat{\mathbf{z}},-\gamma)\mathbf{a}, \tag{20}$$

where $\gamma = \arctan(p_y/p_x)$, and $p_{sy} = 0$, referring to Fig. 3(a.1).

Each boundary of the dexterous workspace is projected as a curve or a line segment on the unit circular area. As shown by later derivations, using this projected unit circular area facilitates the representation of the dexterous workspace from the workspace boundaries.

In the configuration CI-1, there are four type-I boundaries obtained when #1) $\theta_2 = \theta_{2+}$, #2) $\theta_1 = \theta_{1+}$, #3) $L_1 = (r_{1-})\cdot\theta_1$, and #4) $L_1 = L_{1+}$. Note that although $\theta_1$ and $\theta_2$ have a lower limit 0 by definition, this limit does not constrain the dexterity of the end effector. This is because that when the $t^{th}$ segment is straight (namely $\theta_t = 0$), it can bend and change the end effector orientation to all directions. Since $\theta_2$ is the only configuration variable in (19), the boundary #1 corresponds to (19) when $\theta_2 = \theta_{2+}$. It is noted that $a_{sy}$ is multiplied by $p_{sy}$ in (19) and $p_{sy} = 0$ from (20). Hence, $a_{sy}$ disappears from (19). Rearranging (19) in terms of $a_{sx}$ and $a_{sz}$ gives (21). The boundary #1 is hence a straight line on the symmetry plane, corresponding to a circle on the unit spherical surface, as $\theta_2$ set to $\theta_{2+}$ in (21).

$$A_1(\theta_2)a_{sx} + B_1(\theta_2)a_{sz} + C_1(\theta_2) = 0, \tag{21}$$

where $A_1(\theta_2) \triangleq -2p_{sx}(L_{2rz} + L_{2g}\cos\theta_2)$, $B_1(\theta_2) \triangleq -L_{2g}^2$
$+p_{sx}^2 - L_{2rz}^2 - 2L_{2g}L_{2rz}\cos\theta_2$, $\quad C_1(\theta_2) \triangleq 2L_{2g}L_{2rz}$
$+(L_{2g}^2 + p_{sx}^2 + L_{2rz}^2)\cos\theta_2$, and $L_{2rz} = l_2 + L_r + p_{sz}$.

The boundary #2 is obtained by first substituting $l_1$ in (8) with (18). $a_{sy}$ disappears from (18) due to its multiplication with $p_{sy} = 0$. Rearranging (8) in terms of $a_{sx}$ and $a_{sz}$ gives:

$$A_2(\theta_2)a_{sz}^2 + B_2(\theta_{1+},\theta_2)a_{sz} + C_2(\theta_{1+},\theta_2)a_{sx} + D_2(\theta_{1+},\theta_2) = 0, \quad (22)$$

where $A_2(\theta_2) \triangleq -2L_{2g}^2$, $B_2(\theta_1,\theta_2) \triangleq -2L_{2g}L_{2rz}(\cos\theta_1 - 1)$, $C_2(\theta_1,\theta_2) \triangleq -2L_{2g}p_{sx}(1+\cos\theta_1)$, and $D_2(\theta_1,\theta_2) \triangleq L_{2g}^2 - L_{2rz}^2 + p_{sx}^2 + (L_{2g}^2 + L_{2rz}^2 + p_{sx}^2)\cos\theta_1$.

Equation (22) represents a family of curves controlled by $\theta_2$. Since the end effector orientation $^s\mathbf{a}$ also satisfies (21), the boundary #2 is the trajectory of the intersection of the line (21) and the curve (22). Substituting $a_{sx}$ in (22) with (21) gives (23). Given $\theta_2$, $a_{sz}$ and $a_{sx}$ can be solved from (23) and (21) respectively. Therefore, the boundary #2 is a parametric curve driven by $\theta_2$.

$$A_2(\theta_2)a_{sz}^2 + (B_2(\theta_{1+},\theta_2) - C_2(\theta_{1+},\theta_2)B_1(\theta_2)/A_1(\theta_2))a_{sz} \\ + (D_2(\theta_{1+},\theta_2) - C_2(\theta_{1+},\theta_2)C_1(\theta_2)/A_1(\theta_2)) = 0. \quad (23)$$

Next, substituting (9) into (7) yields (24).

$$l_1 = \frac{-L_{2g}^2 - (L_r + l_2)^2 - 2L_{2g}(L_r + l_2)\cos\theta_2 + p_{sx}^2 + p_{sz}^2}{2(L_{2g}\cos\theta_2 + L_{2rz})}. \quad (24)$$

For the boundaries #3 and #4, $\theta_1$ is given by (25) and (26), substituting $L_1$ in (3) with its lower limit $(r_{1-})\cdot\theta_1$, and higher limit $L_{1+}$, respectively.

$$\theta_1 = 2\arctan(l_1/r_{1-}). \quad (25)$$

$$\theta_1/\tan(\theta_1/2) = L_{1+}/l_1. \quad (26)$$

Rearranging (8) in terms of $a_{sx}$ and $a_{sz}$ gives (27).

$$A_3(l_1,\theta_1)a_{sx} + B_3(l_1,\theta_1)a_{sz} + C_3(l_1,\theta_1) = 0, \quad (27)$$

where $A_5(l_1,\theta_1) \triangleq 2p_{sx}(L_{1z} + L_{1gr}\cos\theta_1)$, $B_5(l_1,\theta_1) \triangleq p_{sx}^2 - L_{1gr}^2 - L_{1z}^2 - 2L_{1gr}L_{1z}\cos\theta_1$, $L_{1z} = l_1 - p_{sz}$, $L_{1gr} = l_1 - L_g + L_r$ and $C_5(l_1,\theta_1) \triangleq 2L_{1gr}L_{1z} + (p_{sx}^2 + L_{1gr}^2 + L_{1z}^2)\cos\theta_1$.

Since $l_1$ and $\theta_1$ are expressed as functions of $\theta_2$ using (24), (25) and (26), the straight line (27) is in fact controlled by $\theta_2$. The boundaries #3 and #4 are trajectories of the intersection point of (21) and (27), as parametric curves driven by $\theta_2$. Please note that $\theta_1$ is solved from (26) via a numerical process.

A particular case of $L_1 = (r_{1-})\cdot\theta_1$ is that $\theta_1 = 0$. Then $L_1 = 0$ corresponds to a configuration that has one bending segment, and there exists only one IK solution for a given end effector position. Therefore, $L_1 = 0$ is represented by a dot in the unit circular area rather than a curve.

As for the type-II boundary, an enveloping curve is expected as none of the configuration variables are at their limits. It is noted that (21) gives a set of end effector directions expressed in $\{s\}$ when $\theta_2$ is set to a specific value. Therefore, the set of end effector directions is swept by (21) from $\theta_2 = 0$ to $\theta_2 = \pi$. The type-II boundary is thus the envelope of the family of straight line (21). This boundary is obtained by analytically solving (28), referring to the singularity theory in [18]:

$$A_1 a_{sx} + B_1 a_{sz} + C_1 = 0, \quad \frac{dA_1}{d\theta_2}a_{sx} + \frac{dB_1}{d\theta_2}a_{sz} + \frac{dC_1}{d\theta_2} = 0. \quad (28)$$

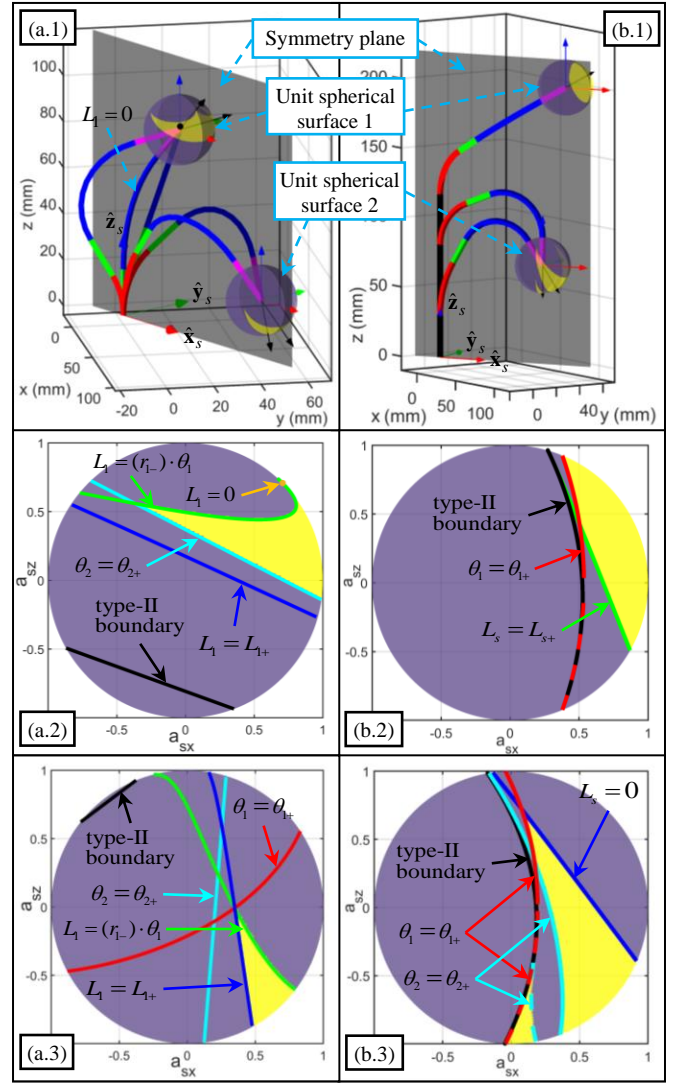

Fig. 3 Dexterous workspace of the continuum robot. The symmetry plane, coordinate $\{s\}$, and the unit spherical surfaces with the continuum robot in CI-1 and CI-2 are shown in (a.1) and (b.1), respectively. The projections of the unit spherical surfaces on the symmetry plane are shown in (a.2) & (a.3) for the configuration CI-1, and (b.2) & (b.3) for the configuration CI-2. The dexterous workspace is represented by the yellow areas.

To determine which side of the boundary curve corresponds to the feasible range of the configuration variable, the configuration limits are perturbed towards the feasible range and substituted into the type-I boundary curves to generate points on the feasible side. The area characterized by the feasible sides of all type-I boundaries and the type-II boundary is the dexterous workspace. The dexterous workspace and its boundaries for the configuration CI-1 are shown in Fig. 3(a.2 and a.3).

B. *Dexterous Workspace of the Configuration CI-2*

In the configuration CI-2, there are four type-I boundaries: #1) $\theta_1 = \theta_{1+}$, #2) $\theta_2 = \theta_{2+}$, #3) $L_s = 0$, and #4) $L_s = L_{s+}$.

For the boundaries #1 and #2, equation (7) is substituted into (8) and (9) to eliminate $L_s$. $a_{sy}$ disappears from (8) and (9) due to its multiplication with $p_{sy} = 0$. Rearranging (8) and (9) in terms of $a_{sx}$ and $a_{sz}$ yields (29) and (30):

$$a_{sx} = \left(A_4(\theta_2)a_{sz}^2 + B_4(\theta_1,\theta_2)a_{sz} + C_4(\theta_1,\theta_2)\right)/p_{sx}, \quad (29)$$

$$A_5(\theta_2)a_{sz}^2 + B_5(\theta_1,\theta_2)a_{sz} + C_5(\theta_1,\theta_2) = 0, \quad (30)$$

where $A_4(\theta_2) \triangleq -L_{2g}$, $B_4(\theta_1,\theta_2) \triangleq -L_{1r2}\cos\theta_1$, and $C_4(\theta_1,\theta_2) \triangleq L_{2g} + L_{1r2}\cos\theta_2$, as well as $A_5(\theta_2) \triangleq L_{2g}^2$, $B_5(\theta_1,\theta_2) \triangleq 2L_{2g}L_{1r2}\cos\theta_1$, and $C_5(\theta_1,\theta_2) \triangleq L_{1r2}^2(\cos^2\theta_1 - 1) - L_{2g}^2 + p_{sx}^2 - 2L_{2g}L_{1r2}\cos\theta_2$.

The boundaries #1 and #2 are expressed in (31) and (32), respectively, by setting $\theta_1$ and $\theta_2$ to their limit values $\theta_{1+}$ and $\theta_{2+}$, using in (29) and (30):

$$\begin{cases} A_5(\theta_2)a_{sz}^2 + B_5(\theta_{1+},\theta_2)a_{sz} + C_5(\theta_{1+},\theta_2) = 0 \\ a_{sx} = \left(A_4(\theta_2)a_{sz}^2 + B_4(\theta_{1+},\theta_2)a_{sz} + C_4(\theta_{1+},\theta_2)\right)/p_{sx} \end{cases}, (31)$$

$$\begin{cases} A_5(\theta_{2+})a_{sz}^2 + B_5(\theta_1,\theta_{2+})a_{sz} + C_5(\theta_1,\theta_{2+}) = 0 \\ a_{sx} = \left(A_4(\theta_{2+})a_{sz}^2 + B_4(\theta_1,\theta_{2+})a_{sz} + C_4(\theta_1,\theta_{2+})\right)/p_{sx} \end{cases}. (32)$$

As for the singularity case of $p_{sx} = 0$, the position of the end effector then lies on $\hat{\mathbf{z}}_w$, and the dexterous workspace is symmetric around $\hat{\mathbf{z}}_w$, which means that the boundaries are horizontal circles parallel to the XY-plane of $\{w\}$.

Please note that the boundaries #1 and #2 are both parabolic and it is possible that two sets of feasible solutions could be solved from (31) and (32), corresponding to two disconnected dexterous workspaces, as shown in Fig. 3(b.3). This means that in such a case, the robot cannot achieve end effector orientations in both dexterous workspaces without changing the end effector position. This very much likely causes the Jacobian-based IK method to fail if the position error is not allowed to increase. The VS-IK method, on the other hand, can find solutions for the target orientations in both dexterous workspaces, by directly solving (13).

According to the configuration transition between CI-1 and CI-2, the boundary #3 in the configuration CI-2 is the same as the boundary #4 in CI-1 ($L_1 = L_{1+}$). And the boundary #4 in CI-2 is obtained by substituting $p_{sz}$ with $p_{sz} - L_{s+}$ in the formulation of the boundary #3 in the configuration CI-2.

It is noted that (29) and (30) give the end effector direction when $\theta_1$ and $\theta_2$ are given. Thus, similar to that in CI-1, the type-II boundary in CI-2 is the envelope of the family of parametric curves (29) and (30). This boundary is obtained numerically by solving (33), referring to [18]:

$$\frac{\partial a_{sx}}{\partial \theta_1}\frac{\partial a_{sz}}{\partial \theta_2} - \frac{\partial a_{sx}}{\partial \theta_2}\frac{\partial a_{sz}}{\partial \theta_1} = 0. \quad (33)$$

The dexterous workspace and its boundaries for the configuration CI-2 are shown in Fig. 3(b.2) and (b.3).

C. *Unreachable Target Poses*

For a target pose with the target position is within the translational workspace, the proposed VS-IK method would try to solve (13) and (19). If the VS-IK method gives a real number solution that violates the configuration limits, the target direction lies on the feasible side of the type-II boundary. Therefore, the point closest to the target direction lies on the type-I boundaries. Such closest points are found by discretizing the type-I boundaries as specifying sequences of $\theta_1$ and $\theta_2$, and analytically calculating $a_{sx}$ and $a_{sz}$. If the VS-IK method gives a non-real solution or fails to converge, the target direction is infeasible since it violates the type-II boundary of the continuum robot. The closest feasible direction then lies on the type-II boundary. The type-II boundary can be solved from (28) for CI-1 and (33) for CI-2.

The proposed formulation has several advantages over the existing numerical methods. The existing methods only give an approximation to the dexterous workspace by discretizing the end-effector orientation into patches on the unit spherical surface. Also, these methods identify the dexterous workspace either by solving the IK for every patch with the Jacobian-based method [14] or by generating feasible end-effector poses with the Monte-Carlo method [15, 16], both of which are time-consuming. The proposed formulation, in contrast, gives exact boundaries that can be efficiently calculated.

V. SIMULATION STUDY

Numerical simulation was conducted to evaluate the performance of the proposed VS-IK method. The Jacobian-based damped least squares method [19] (the Jacobian-DLS method) was used for comparison.

The IK methods were considered to be solved when the position error and the orientation error of the end effector was less than 0.01 mm and 0.01 rad respectively, without exceeding the limits of the configuration variables and the number of iterations. The orientation error was calculated as the rotation angle between the target orientation and the end effector orientation.

The VS-IK method employed the Newton-Raphson method to solve the nonlinear equations (13) and (19). The iteration residuals of (13) and (19) were empirically adjusted to 0.01 and 0.0003, respectively, to match the designated pose error tolerances. The maximum iterations per target pose was set to 200 for both methods. The initial configurations for the Jacobian-DLS method were set to $L_1 = L_{1+}/2$ for CI-1 and $L_s = L_{s+}/2$ for CI-2, with all other configuration variables equal to zero. For the VS-IK method, the initial values were set to $\theta_2 = \theta_{2+}/2$ for CI-1 and $\theta_1 = \arccos(|a_z|)$ for CI-2. Both methods were implemented in MATLAB (Mathworks Inc.) and ran on a laptop with a 2.3 GHz Intel Core i5-8300H CPU. The structural parameters of the continuum robot used in the simulation are given in Table II.

A. *Case Study for Reachable Targets*

This case study was conducted for 250,000 test cases (sets of initial and target poses for verifying the effectiveness of the IK approach) from a previous work [17] for comparison (namely, 125,000 cases for Configuration #3 in [17] as configuration CI-1 in this study, as well as 125,000 cases for Configuration #4 in [17] as configuration CI-2 in this study). The target poses were generated by assigning random values to the configuration variables within the feasible ranges and calculating forward kinematics, and hence are reachable.

The performance of the two IK methods is shown in Table III. The VS-IK method used substantially less computation time and the average time for each iteration, compared to the Jacobian-DLS method. For the configurations CI-1 and CI-2, the total computation time of the VS-IK method for 125,000 cases was 2.86 s and 7.13 s, respectively, corresponding to 99.43% and 96.72% time reduction compared to the Jacobian-DLS method, respectively. In terms of the time per iteration, the VS-IK method is more than 10 times faster than the Jacobian-based method, referring to Table III, facilitating real-time trajectory tracking tasks. The major computational load of the VS-IK method only comes from computing the values and gradients of one-variable equations (13) and (19). On the other hand, the Jacobian-DLS method calculates the forward kinematics and the pseudo-inverse of the Jacobian matrix at every iteration.

The VS-IK method achieved 100% success rate. The total success rate of the Jacobian-DLS method in this study is higher than that in our previous work (97.79%) because the configuration transition is not considered in this study. Nevertheless, the convergence of the Jacobian-DLS method relied on the proper setting of the initial configuration of the robot and the end effector twist (i.e., the linear and the angular velocities). In contrast, all targets in the conducted simulation were successfully converged from a single initial value setting using the VS-IK method, without tuning extra parameters.

### B. Case Study for Non-Guaranteed Reachable Targets

This case study was conducted on test cases generated by randomly sampling positions within the translational workspace that is generated according to [14], and generating uniformly distributed random Euler angles to obtain random orientations in $SO(3)$. Therefore, all target positions were reachable within the configuration limits, but the corresponding orientations were not guaranteed to be reachable. This case study emulates a realistic teleoperation scenario where the target poses might not be completely reachable. Two test sets were generated for CI-1 and CI-2, respectively, each containing 125,000 poses.

The dimension reduced Jacobian formulation [20] was used as this Jacobian-DLS method with position reaching as the primary task [17] to converge to target poses that may be out of the dexterous workspace. Since the orientation might not converge successfully, the iteration was set to terminate if the orientation error was not smaller than the previous smallest error for 30 consecutive iterations after converging to the target position. For the VS-IK method, the iteration terminated if the configuration variables $\theta_1$ and $\theta_2$ exceeded $\pi$ for 3 consecutive iterations. If the target orientation was out of the dexterous workspace, 600 equally incremental values of $\theta_1$ and $\theta_2$ within their limits were used to generate the boundaries.

Table IV shows the performance of both IK methods. Similar to the simulation in Section V-A, the VS-IK method exhibited improved computational efficiency and 100% success rate, achieving smaller orientation error compared to the Jacobian-DLS method.

The success rate of the Jacobian-DLS method was significantly lower than that of the VS-IK method. Representative failed cases are shown in Fig. 4 (a), where the Jacobian-DLS method was stuck at the configuration variables limits while trying to reach the lower part of the translational workspace. In the CI-1 case, the Jacobian-DLS method first retracted the 1st segment to reduce the position error, as in Fig. 4(d). In the CI-2 case, the Jacobian-DLS method first retracted the base segment to reduce the position error, as in Fig. 4(e). To reach the target position, the robot had to extend and bend the 1st segment in the CI-1 case and bend the 1st segment in the CI-2 case. However, doing so would temporarily increase the position error in both cases, which is prohibited since the Jacobian-DLS method is gradient-based in the task space. Therefore, the Jacobian-DLS method was stuck at $L_1 = 0$ in the CI-1 case and $L_s = 0$ in the CI-2 case. In contrast, the VS-IK method can successfully reach the target positions while returning the closest orientation in the dexterous workspace to the target orientation, as shown in Fig. 4(b) and 4(c).

TABLE II. Structural Parameters

| $L_{10}$ | $L_{20}$ | $L_r$ | $L_g$ | $L_{s+}$ | $\theta_{1+}$ | $\theta_{2+}$ | $r_{1-}$ |
|---|---|---|---|---|---|---|---|
| 40 mm | 60 mm | 20 mm | 20 mm | 150 mm | $\pi/2$ rad | $2\pi/3$ rad | $80/\pi$ mm |

TABLE III. Performance Comparison for Reachable Targets

| Performance Index | Jacobian-DLS | | VS-IK | |
|---|---|---|---|---|
| | CI-1 | CI-2 | CI-1 | CI-2 |
| Computation time (s) | 464.18 | 217.40 | 2.86 | 7.13 |
| Avg. iterations | 26.73 | 14.59 | 3.86 | 7.05 |
| Avg. time per iteration ($10^{-5}$ s) | 13.89 | 11.92 | 0.59 | 0.81 |
| Success rate | 97.58% | 99.43% | **100%** | **100%** |
| Failure count | 3025 | 715 | 0 | 0 |

TABLE IV. Performance Comparison for Non-guaranteed Reachable Targets

| Performance Index | Jacobian-DLS | | VS-IK | |
|---|---|---|---|---|
| | CI-1 | CI-2 | CI-1 | CI-2 |
| Execution time (s) | 2218.02 | 2030.56 | 18.34 | 60.43 |
| Avg. iterations | 72.48 | 84.07 | 3.47 | 14.75 |
| Success rate | 64.89% | 90.19% | **100%** | **100%** |
| Failure count | 43889 | 12268 | 0 | 0 |
| Avg. position error (mm) | 11.73 | 4.92 | 0.00073 | 0.0011 |
| Avg. orientation error (rad) | 1.1293 | 0.6128 | 0.9521 | 0.5980 |

## VI. Conclusions

In this paper, an efficient IK method for continuum robots with one or two inextensible bending segments is presented. By using the line segment representation to separate the configuration variables, the IK problem is formulated as solving a one-dimensional nonlinear equation for $\theta_1$ or $\theta_2$, and then solving other configuration variables in closed-form. By incorporating the configuration limits in the formulation of the VS-IK, the boundaries of the dexterous workspaces are formulated as parametric curves. A comparative simulation study was performed for the VS-IK method and the Jacobian-DLS method on a total of 500,000 test cases. The results showed that the VS-IK method achieved more than 96% computation time reduction and 100% success rate. The efficiency of the VS-IK method would facilitate real-time motion planning and control. The analytic formulation of the dexterous workspaces can also benefit the design and kinematics performance evaluation of continuum robots.

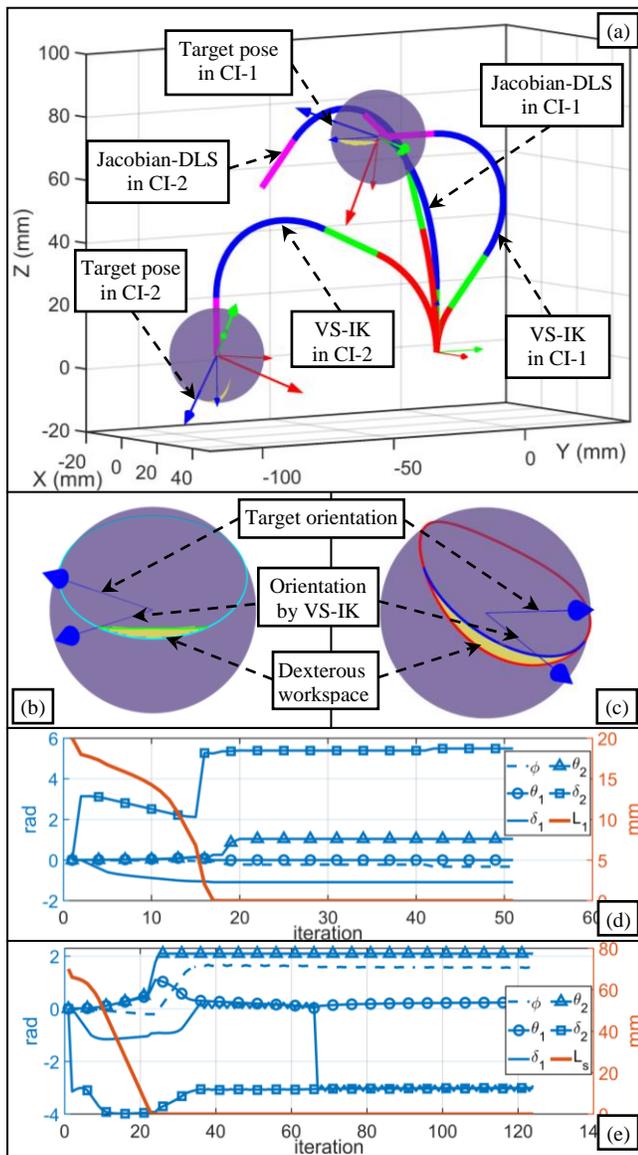

Fig. 4 Representative failed cases of the Jacobian-DLS method and the solutions given by the VS-IK method are shown in (a). Target orientations, orientations given by the VS-IK method, and the dexterous workspace at the target positions are shown in (b) for the CI-1 case and (c) for the CI-2 case. The trajectories of the configuration variables of the Jacobian-DLS method are shown in (d) for the CI-1 case and (e) for the CI-2 case.

Extending the VS-IK method to continuum robots with more than two segments will be attempted in the future work.